\documentclass[a4paper,fleqn]{cas-dc}
\usepackage[numbers,sort&compress]{natbib}
\usepackage{amsmath,amssymb,booktabs,graphicx,tabularx,xcolor,placeins}
\AddToHook{env/Abstract/begin}{}
\allowdisplaybreaks[0]

\newcommand{\relu}{\operatorname{ReLU}}
\newcommand{\crelu}{\operatorname{CReLU}}
\newcommand{\R}{\mathbb{R}}
\newcommand{\C}{\mathbb{C}}
\graphicspath{{figures/}}
\begin{document}
\let\WriteBookmarks\relax
\renewcommand{\topfraction}{0.95}
\renewcommand{\dbltopfraction}{0.95}
\renewcommand{\textfraction}{0.05}
\renewcommand{\floatpagefraction}{0.65}
\renewcommand{\dblfloatpagefraction}{0.65}
\shorttitle{Learning Nonlinear Interactions with PURe}
\shortauthors{Li et al.}
\title[mode=title]{Learning nonlinear interactions and cross-feature couplings with product-unit residual networks}
\author[1,2]{Ziyuan Li}
\cormark[1]\ead{ziyuan.li@tum.de}
\author[1]{Uwe Jaekel}
\author[1]{Babette Dellen}
\affiliation[1]{organization={Department of Mathematics, Informatics and Technology,
University of Applied Sciences Koblenz},addressline={Joseph-Rovan-Allee 2},
postcode={53424},city={Remagen},country={Germany}}
\affiliation[2]{organization={Technical University of Munich},city={Munich},country={Germany}}
\cortext[1]{Corresponding author}

\begin{abstract}
Understanding nonlinear feature interactions is crucial in science and engineering, yet standard multilayer perceptrons (MLPs) often capture such interactions only implicitly, leading to entangled representations that can impair robustness and interpretability. We investigate complex-valued product-unit residual networks (CV-PURe) that integrate multiplicative product units with residual connections to explicitly model cross-feature couplings while stabilizing optimization. We conduct a systematic evaluation on an interaction-driven synthetic benchmark and real-world datasets and evaluate prediction together with input-level interaction recovery under a reproducible protocol. Across five data splits and five training seeds, CV-PURe achieves the lowest mean RMSE on four of six scientific benchmarks, reducing Friedman error by 53.9\% relative to MLP. Its Friedman interaction maps assign 60.7\% of total strength to the known interacting pair, compared with 31.0\% for MLP. On a synthetic task containing third- and fourth-order products and 20 distractors, CV-PURe reduces RMSE from 0.1376 to 0.0797 relative to a complex-valued MLP and recovers all nine true pairwise projections in every training/sampling repetition. Bike Sharing experiments connect improved prediction to the contribution of temporal, calendar, and seasonal context through paired ablations of both PURe variants. Parkinsons Telemonitoring extends the context analysis to repeated measurements. Noise and learning-curve evaluations demonstrate sustained predictive gains under perturbed inputs and limited supervision, while inference measurements quantify computational cost. This article extends our ICCS 2026 paper, \emph{Modeling Nonlinear Feature Interactions with Product-Unit Residual Networks}~\cite{li2026modeling}.
\end{abstract}

\maketitle

\section{Introduction}\label{sec:introduction}

Modeling nonlinear relationships between input variables is a fundamental challenge occurring in scientific and engineering regression tasks, including physical system modeling, materials science, and socio-economic prediction problems. Often the target variable is not determined solely by individual features, but by nonlinear interactions among multiple factors. Accurately capturing these nonlinear feature interactions is therefore essential for achieving reliable predictions and meaningful scientific insights.

Multilayer perceptron networks (MLP) have become a standard tool for nonlinear regression due to their universal approximation capability and flexible function representation~\cite{cybenko1989approximation,rumelhart1986learning}. Although they can learn nonlinear interactions, conventional MLPs do not explicitly parameterize feature products. Their learned representations can also be difficult to interpret, making it challenging to distinguish meaningful cross-feature dependencies from spurious correlations~\cite{molnar2020interpretable,lundberg2017unified}.

To address these limitations, multiplicative neural architectures based on product units have been proposed as an alternative modeling paradigm~\cite{li2025advancing,li2025pure,dellen2024predicting,li2024data,dellen2019function,leerink1994learning,durbin1989product}. By explicitly encoding multiplicative relationships, product-unit networks provide a natural inductive bias for modeling nonlinear feature interactions that commonly arise in scientific data. Their optimization can nevertheless be sensitive to initialization and to the amplification or attenuation introduced by products~\cite{leerink1994learning,dellen2019function}.

Residual connections have been shown to substantially improve the optimization and stability of deep neural networks by facilitating gradient flow and reducing vanishing gradient effects \cite{he2016deep}. They have been successfully applied to a wide range of learning problems, including regression tasks with complex nonlinear structures \cite{veit2016residual,allen2019convergence}. Recently, the use of residual connections in conjunction with multiplicative product-unit networks has been proposed \cite{li2025pure,li2025advancing,li2026pu}. Although the results have been encouraging, it is unclear how residual connections influence predictive performance, generalization behavior, and interpretability when targeting explicit interaction modeling.

In this work, we study product-unit residual networks (PURe), a class of neural architectures that integrate multiplicative product units with residual connections to explicitly model nonlinear feature interactions. We evaluate predictive accuracy, robustness to noise, and performance under limited training data on synthetic and real-world regression datasets. Our results demonstrate that CV-PURe achieves competitive performance, enhanced generalization, and improved interpretability, making it a promising modeling framework for computational-science applications governed by complex nonlinear relationships.

The main contributions are:
a) a signed-output complex product-unit residual architecture and a reproducible comparison of five neural models across six scientific datasets, five data splits, and five training seeds;
b) a quantitative connection between predictive gains and interaction recovery, combining known-function tests, repeated interaction maps, and a closely parameter-matched complex-valued MLP control; and
c) evidence of effective context use in scientific regression, with predictive comparisons, input-level interaction maps, and paired context ablations of both real- and complex-valued PURe.

This article extends our ICCS 2026 study~\cite{li2026modeling} with a trainable real-valued output projection, cross-split evaluation, ground-truth interaction recovery, and results for context-dependent data (Bike Sharing, Parkinsons Telemonitoring). Noise, learning-curve, and inference experiments establish the practical accuracy and cost of the resulting models.

\section{Related work}\label{sec:related-work}
Product-unit neural networks were introduced to explicitly encode multiplicative interactions between input features~\cite{durbin1989product}. Early work studied both their ability to learn higher-order combinations and the associated optimization difficulties~\cite{leerink1994learning}. Recent applications include function extrapolation~\cite{dellen2019function}, 3D shape completion~\cite{li2024data}, nuclear-mass prediction~\cite{dellen2024predicting,li2026product}, and medical imaging~\cite{li2025advancing,li2026pu}. The aforementioned product-unit networks differ in number of layers, layer structure, loss functions, and the use of real- versus complex-valued weights. To accommodate signed inputs and to enable joint modeling of magnitude and phase information, product-unit representations have been extended to the complex domain~\cite{dellen2024predicting,li2025advancing}. When adding layers and entering deep learning \cite{li2025pure}, residual connections \cite{he2016deep} help stabilize the optimization process. The network PURe embeds the product-unit transformation in the residual branch of the network ~\cite{li2025pure,li2026modeling}.

Other approaches to multiplication use neural mechanisms such as gating and attention~\cite{jayakumar2020multiplicative} or complex affine layers, which instead couple real and imaginary components through weighted sums, with applications in audio processing~\cite{trabelsi2017deep} and MRI tissue-parameter estimation~\cite{virtue2017better}.

Explicit feature crossing provides another way to introduce multiplication. Factorization machines model pairwise interactions using factorized coefficients~\cite{rendle2010factorization}, while the cross branch of DCN V2 constructs bounded-degree polynomial interactions alongside a deep network~\cite{wang2021dcn}. More closely related to product units, the Adaptive Factorization Network learns exponents over positive feature embeddings through a logarithmic transformation layer~\cite{cheng2020adaptive}.

Interpretable architectures expose main effects and pairwise interactions, as in GAMI-Net~\cite{yang2021gami}, or shared predictive features across outputs, as in Common Structure Neural Networks~\cite{gu2026interpretable}. For fitted predictors, SHapley Additive exPlanations (SHAP) attributes predictions to individual features~\cite{lundberg2017unified}. Joint effects are examined through rule-ensemble interaction strength~\cite{friedman2008predictive}, weight-based Neural Interaction Detection~\cite{tsang2017detecting}, Integrated Hessians~\cite{janizek2021explaining}, Shapley--Taylor indices~\cite{sundararajan2020shapley}, and perturbation-based Archipelago~\cite{tsang2020does}.

\section{Method}\label{sec:method}
\subsection{Product units}
Given an input vector $\mathbf{x}=(x_1,\dots,x_d)$, a classical product unit computes
\begin{equation}
y=\prod_{i=1}^{d}x_i^{w_i},
\label{eq:product}
\end{equation}
where $w_i$ denotes a learnable exponent associated with the $i$-th input \cite{durbin1989product, dellen2019function}. This formulation enables compact representations of high-order feature interactions with products, powers, and ratios directly available. The weights can be real- or complex values, depending on the architectural choice. 

For numerical stability, product units are commonly implemented using a logarithmic reparameterization~\cite{dellen2019function}:
\begin{equation}
y=\exp\!\left(\sum_{i=1}^{d}w_i\log x_i\right).
\label{eq:log-product}
\end{equation}
This log-linear-exponential form highlights the multiplicative inductive bias of product units and distinguishes them structurally from additive neural representations.

\subsection{Product-unit residual block}
Let $\mathbf{h}\in\R^h$ denote the input of a residual block, where $h$ is the hidden dimension. Our real-valued product-unit residual block is implemented as a three-stage transformation followed by an identity skip connection:
\begin{equation}
\mathbf{h}_{\mathrm{out}}=\mathbf{h}
+\mathrm{FC}_2\!\left(\mathrm{PU}\!\left(\mathrm{FC}_1(\mathbf{h})\right)\right).
 \label{eq:real-block}
\end{equation}
Here, $\mathrm{FC}_1$ and $\mathrm{FC}_2$ are affine mappings (linear layers with bias) from $\R^h$ to $\R^h$.

Given an intermediate vector $\mathbf{a}=\mathrm{FC}_1(\mathbf{h})\in\R^h$, the product-unit mapping is computed using a positivity-enforced preprocessing:
\begin{equation}
\mathbf{s}=\relu(\mathbf{a})+\mathbf{1},\qquad
\relu(\cdot)=\max(\cdot,0),
\label{eq:real-shift}
\end{equation}
where $\mathbf{1}\in\R^h$ denotes the all-ones vector, followed by a log-linear-exponential computation:
\begin{equation}
\mathrm{PU}(\mathbf{a})=\exp\!\left(W_{\mathrm{pu}}\log\mathbf{s}\right).
\label{eq:real-pu}
\end{equation}
Here $W_{\mathrm{pu}}\in\R^{h\times h}$ is a learnable weight matrix (without bias), and $\log(\cdot)$ and $\exp(\cdot)$ are applied element-wise. The shift ensures positive factors and a well-defined logarithm. When $a_k\leq0$, $s_k=1$ is multiplicatively neutral; positive activations produce factors greater than one.

\begin{figure}[pos=t]
\centering
\includegraphics[width=0.85\linewidth]{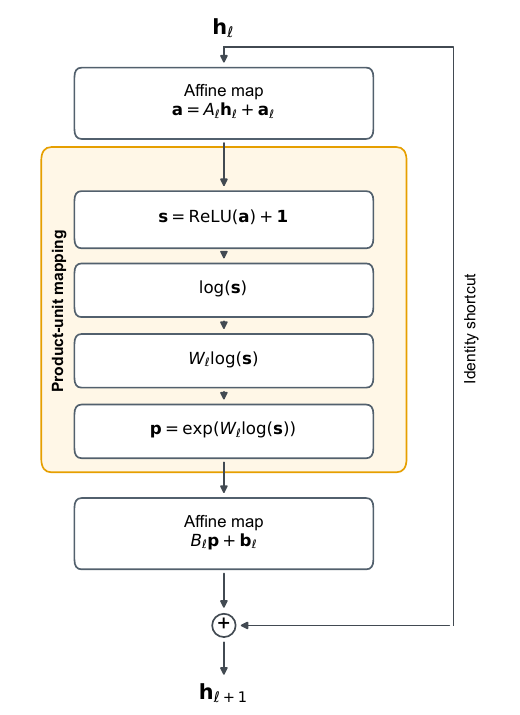}
\caption{Real-valued product-unit residual block. $A_\ell$, $B_\ell$, and $W_\ell$ denote the weights of $\mathrm{FC}_1$, $\mathrm{FC}_2$, and the product-unit layer.}
\label{fig:pure-block}
\end{figure}

The affine maps learn the factors and return their products to the residual state (Fig.~\ref{fig:pure-block}).

\subsection{Overall real-valued network}
\begin{figure}[pos=t]
\centering
\includegraphics[width=0.92\linewidth]{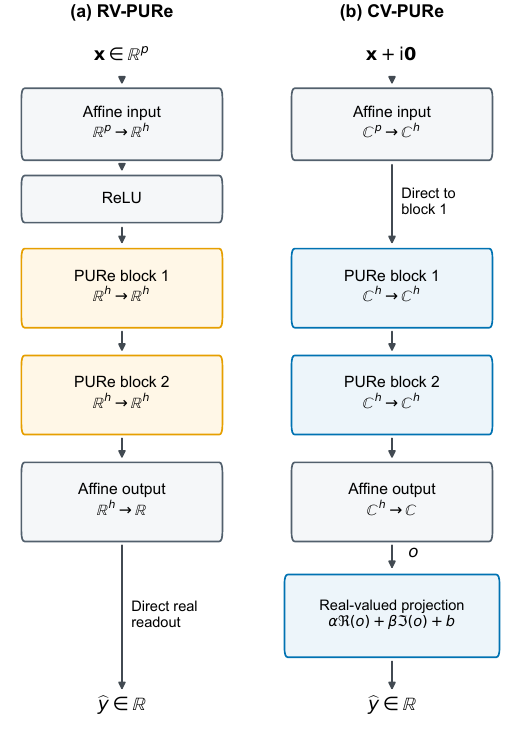}
\caption{Overall architectures of (a) RV-PURe and (b) CV-PURe. The CV-PURe shown here uses a trainable real-valued projection to produce signed predictions.}
\label{fig:pure-architectures}
\end{figure}

Given an input $\mathbf{x}\in\R^p$, the network first maps it to the hidden space and applies an element-wise rectified linear unit:
\begin{equation}
\mathbf{h}^{(0)}=\relu\!\left(\mathrm{FC}_{\mathrm{in}}(\mathbf{x})\right).
\end{equation}
Then, two residual blocks are applied sequentially:
\begin{equation}
\mathbf{h}^{(1)}=\mathcal{B}_1\!\left(\mathbf{h}^{(0)}\right),\qquad
\mathbf{h}^{(2)}=\mathcal{B}_2\!\left(\mathbf{h}^{(1)}\right).
\end{equation}
Here, $\mathcal{B}_1$ and $\mathcal{B}_2$ denote the block in Eq.~\eqref{eq:real-block}.
Finally, the prediction is produced by a linear output layer:
\begin{equation}
\widehat{y}=\mathrm{FC}_{\mathrm{out}}\!\left(\mathbf{h}^{(2)}\right).
\end{equation}
Figure~\ref{fig:pure-architectures}(a) shows the complete network.

\subsection{Complex-valued product-unit residual block}
To accommodate signed inputs and enable joint modeling of magnitude and phase information, we extend the block to the complex domain. Real inputs are embedded as $\mathbf{x}+\mathrm{i}\mathbf{0}$.
Let $\mathbf{z}\in\C^h$ be the input of a complex-valued block. The block structure mirrors the real-valued case:
\begin{equation}
\mathbf{z}_{\mathrm{out}}=\mathbf{z}
+\mathrm{FC}_2^{\C}\!\left(\mathrm{PU}_{\C}\!\left(\mathrm{FC}_1^{\C}(\mathbf{z})\right)\right).
\label{eq:complex-block}
\end{equation}
Given $\mathbf{a}=\mathrm{FC}_1^{\C}(\mathbf{z})$, we apply preprocessing by separately rectifying the real and imaginary parts and adding a constant offset:
\begin{align}
 \widetilde{\mathbf{a}}={}&
 \bigl(\relu(\Re\mathbf{a})+c\mathbf{1}\bigr)\nonumber\\
 &+\mathrm{i}\bigl(\relu(\Im\mathbf{a})+c\mathbf{1}\bigr),
 \qquad c=2^{-1/2}.
 \label{eq:complex-shift}
\end{align}
The complex product-unit transformation is
\begin{equation}
 \mathrm{PU}_{\C}(\mathbf{a})
 =\exp\!\left(W_{\mathrm{pu}}^{\C}\operatorname{Log}
 \widetilde{\mathbf{a}}\right),
 \label{eq:complex-pu}
\end{equation}
where $W_{\mathrm{pu}}^{\C}\in\C^{h\times h}$ is a complex-valued weight matrix (without bias), and $\operatorname{Log}(\cdot)$ and $\exp(\cdot)$ denote the complex logarithm and exponential applied element-wise. We use the principal logarithm. Every component of $\widetilde{\mathbf{a}}$ lies in the first quadrant and has modulus at least one.

The conference formulation used the magnitude of the complex scalar output $o$ as its network-scale prediction, $\widehat{y}=|o|$~\cite{li2026modeling}, which restricts this prediction to nonnegative values.

Let $\mathbf{r}=|\widetilde{\mathbf{a}}|$ and $\boldsymbol{\phi}=\operatorname{Arg}\widetilde{\mathbf{a}}$ denote element-wise modulus and principal phase. Writing $\mathbf{p}_{\C}=\mathrm{PU}_{\C}(\mathbf{a})$, $W_{\mathrm{pu}}^{\C}=U+\mathrm{i}V$, and
$\operatorname{Log}\widetilde{\mathbf{a}}=\log\mathbf{r}+\mathrm{i}\boldsymbol{\phi}$ gives
\begin{align}
 \mathbf{q}&=U\log\mathbf{r}-V\boldsymbol{\phi},\qquad
 \boldsymbol{\omega}=V\log\mathbf{r}+U\boldsymbol{\phi},\nonumber\\
 \mathbf{p}_{\C}&=\exp(\mathbf{q})\odot
 \left[\cos\boldsymbol{\omega}+\mathrm{i}\sin\boldsymbol{\omega}\right].
 \label{eq:amplitude-phase}
\end{align}
Here $\odot$ denotes element-wise multiplication. The learned exponents couple log-amplitude and phase, producing both amplitude modulation and oscillatory transformations.

In CV-PURe (Fig.~\ref{fig:pure-architectures}(b)), a complex affine input layer feeds the first of two complex residual blocks without an intervening ReLU. A complex affine output layer produces the scalar $o$. All complex-valued models in this article use a trainable real-valued projection:
\begin{equation}
 \widehat{y}=\alpha\Re(o)+\beta\Im(o)+b,
 \qquad \alpha,\beta,b\in\R.
 \label{eq:projection}
\end{equation}
The three readout parameters are learned jointly with the network. This affine projection allows both positive and negative predictions, accommodating signed responses and centered, standardized targets. CV-MLP uses the same projection.

\subsection{Input-level interaction analysis}
\label{sec:interaction-method}
Product units multiply learned latent features, whereas interpretation concerns relationships between the original inputs. We apply the same mixed finite-difference score to the complete input--output function of every model.

We use the four-point contrast employed in perturbation-based interaction detection~\cite{tsang2020does}. For an input $\mathbf{x}$ and a reference $\mathbf{r}$, let $\mathbf{x}^{i\leftarrow r_i}$ denote replacement of feature $i$ by its reference value. The mixed finite difference is
\begin{align}
\Delta_{ij}f(\mathbf{x},\mathbf{r})={}&f(\mathbf{x})
-f(\mathbf{x}^{i\leftarrow r_i})
-f(\mathbf{x}^{j\leftarrow r_j})\nonumber\\
&+f(\mathbf{x}^{i,j\leftarrow r_i,r_j}).
\label{eq:finite-difference}
\end{align}
For a function additive in $i$ and $j$, this contrast vanishes. Interaction strength is estimated over $M$ input--reference pairs as the mean absolute contrast,
\begin{equation}
S_{ij}=\frac{1}{M}\sum_{a=1}^{M}
\left|\Delta_{ij}f(\mathbf{x}_a,\mathbf{r}_a)\right|.
\label{eq:interaction-score}
\end{equation}
The predictor $f$ returns standardized targets, making interaction strengths comparable across models within a dataset.

\section{Predictive performance and architectural analysis}\label{sec:protocol}
We evaluate the proposed PURe on six regression benchmarks, covering synthetic interaction-driven data, small-scale real-world engineering data, and large-scale socio-economic data. Cross-split predictive comparisons are followed by a paired residual-connection ablation on Concrete.

\subsection{Experimental setup}
\label{sec:shared-protocol}
Real-valued and complex-valued models are evaluated under shared data partitions, optimizer settings, and checkpoint selection within each comparison. All reported experiments use the current architectures, with the projection in Eq.~\eqref{eq:projection} for CV-PURe and CV-MLP. Inputs and targets are standardized using training data only. MSE is measured on the standardized target scale; MAE and RMSE are obtained after inverse transformation to the original target units.

\paragraph{Architectures.}\label{sec:architectures}
The real MLP contains seven hidden affine layers with ReLU activations and a scalar linear output. ResMLP uses an input ReLU layer and two additive residual blocks; each branch contains three affine layers with ReLU between successive layers. A final ReLU precedes its scalar output. CV-MLP has seven complex hidden affine layers with $\crelu(z)=\relu(\Re z)+\mathrm{i}\relu(\Im z)$, followed by a complex scalar output and the real-valued projection. On the six scientific benchmarks, real networks use width 128 and complex networks width 64. Synthetic recovery and context experiments use width 64 for all neural models. CV-MLP provides an additive complex-valued control with nearly the same parameter count as CV-PURe.

\paragraph{Optimization and initialization.}
All models are trained using the mean squared error loss, i.e.,
\begin{equation}
\mathcal{L}=\frac{1}{N}\sum_{i=1}^{N}(y_i-\widehat{y}_i)^2,
\label{eq:mse-loss}
\end{equation}
where $y_i$ and $\widehat{y}_i$ denote the ground-truth target and model prediction, on the training target scale, respectively~\cite{bishop2006pattern}. We use the Adam optimizer~\cite{kingma2014adam} with learning rate $10^{-3}$. Scientific benchmarks use at most 200 epochs, batch size 128, and early-stopping patience 20. Synthetic recovery and context tasks use at most 100 epochs, batch size 256, and patience 15. For evaluation on the test set, the model checkpoint achieving the lowest validation loss is selected.

Affine layers use PyTorch's default initialization, with the exceptions specified below. The two PURe residual blocks are initialized independently; only the final output affine layer is initialized to zero. The projection starts at $(\alpha,\beta,b)=(1,0,0)$. ResMLP initializes the last affine map of its second residual branch and its final output layer to zero. PURe variants use gradient-norm clipping at 1.0, also retained in the shortcut and context ablations. All models are trained on uncorrupted inputs.

\paragraph{Seeds and computational environment.}
We separate the sources of randomness. Data-generation seed 0 fixes Friedman and the two synthetic response datasets. Split seeds 0--4 define five train/validation/test partitions for each scientific benchmark; each partition uses training seeds 0--4. All other random-partition analyses use split seed 0. Bike Sharing and Parkinsons use the chronological partitions in Table~\ref{tab:datasets}. Training seeds are 0--4, except for the synthetic recovery tasks and Bike context ablations, which use 0--2. Each training seed controls Python, NumPy, PyTorch initialization, and shuffled minibatches. Nested training subsets use selection seed 0. Interaction sampling and noise realizations use seeds 0--4, except that context maps use sampling seed 0. These evaluation samples are shared across models.

\begin{table*}[t]
\caption{Datasets and actual partition sizes. $p$ is the encoded input dimension, including full context where applicable. Random partitions use split seed 0 for the fixed-split analyses; the six scientific benchmarks also use split seeds 1--4.}
\label{tab:datasets}\centering\small
\setlength{\tabcolsep}{10pt}
\begin{tabular}{@{}lrrll@{}}
\toprule
Dataset & Observations & $p$ & Train / validation / test & Partition \\ \midrule
Friedman & 10,000 & 10 & 7,000 / 1,499 / 1,501 & Random \\
Concrete & 1,030 & 8 & 618 / 206 / 206 & Random \\
California Housing & 20,640 & 8 & 16,512 / 2,064 / 2,064 & Random \\
Energy Efficiency & 768 & 8 & 537 / 115 / 116 & Random \\
Yacht Hydrodynamics & 308 & 6 & 215 / 46 / 47 & Random \\
Auto MPG & 392 & 9 & 274 / 58 / 60 & Random \\
Synthetic pairwise & 10,000 & 24 & 7,000 / 1,499 / 1,501 & Random \\
Synthetic higher-order & 10,000 & 27 & 7,000 / 1,499 / 1,501 & Random \\
Bike Sharing & 17,379 & 26 & 10,427 / 3,476 / 3,476 & Chronological \\
Parkinsons & 5,875 & 61 & 3,573 / 1,221 / 1,081 & Within-subject chronological \\
\bottomrule
\end{tabular}
\end{table*}

All fits use the same local environment: PyTorch 2.14.0 with CUDA 13.0 on an NVIDIA GeForce RTX 5080, float32 real operations and complex64 complex operations. Deterministic algorithms are enabled, TF32 is disabled, and float32 matrix multiplication uses the highest-precision setting. Saved configurations record split indices, preprocessing statistics, source hashes, and checkpoint identities.

\paragraph{Statistical summaries.}\label{sec:statistics}
Results are reported as the mean and sample standard deviation across runs. For the cross-split benchmark, we first average the five training runs within each split and then report the mean and sample SD of the five split means. Other prediction and error-curve summaries use training seeds on the fixed partition. Interaction summaries specify their aggregation in each caption; synthetic recovery ranks summarize all training/sampling combinations. Positive paired improvements denote baseline error minus PURe error for matching training seeds; context-ablation effects denote ablated minus full-context error. We use 10,000 seed-level bootstrap resamples, with bootstrap seed 0, for 95\% percentile confidence intervals~\cite{tibshirani1993introduction}. Repeated noise and interaction evaluations are averaged within checkpoints before seed-level inference.

\begin{table*}[pos=tp]
\caption{Cross-split test RMSE in original target units. Each of five split seeds uses five training seeds. Values are the mean and sample SD of the five split-level means. California values are in units of \$100,000.}
\label{tab:scientific}\centering\small
\setlength{\tabcolsep}{10pt}
\begin{tabular}{@{}lrrrrr@{}}
\toprule
Dataset & MLP & ResMLP & RV-PURe & CV-MLP & CV-PURe \\ \midrule
Friedman & $0.1695\pm0.0071$ & $0.1738\pm0.0043$ & $0.1589\pm0.0044$ & $0.1858\pm0.0063$ & $\mathbf{0.0781\pm0.0068}$ \\
Concrete & $5.4253\pm0.3328$ & $\mathbf{5.3830\pm0.3879}$ & $5.4670\pm0.3809$ & $5.5636\pm0.4033$ & $5.5847\pm0.3017$ \\
California & $0.5182\pm0.0101$ & $0.5167\pm0.0101$ & $0.5195\pm0.0107$ & $0.5221\pm0.0099$ & $\mathbf{0.5145\pm0.0093}$ \\
Energy & $0.5922\pm0.0485$ & $0.5758\pm0.0519$ & $0.5785\pm0.0412$ & $0.6180\pm0.0444$ & $\mathbf{0.5115\pm0.0480}$ \\
Yacht & $\mathbf{0.9858\pm0.2169}$ & $1.1165\pm0.2975$ & $1.0559\pm0.1652$ & $1.0959\pm0.2752$ & $1.0849\pm0.3306$ \\
Auto MPG & $2.8517\pm0.5931$ & $2.7626\pm0.5992$ & $2.7972\pm0.5886$ & $2.8074\pm0.6146$ & $\mathbf{2.7198\pm0.4964}$ \\
\bottomrule
\end{tabular}
\end{table*}

\subsection{Benchmark predictive performance}
\label{sec:scientific-benchmarks}
Table~\ref{tab:scientific} evaluates clean-input prediction across data partitions, jointly measuring predictive accuracy and its dependence on the train/validation/test split. Section~\ref{sec:noise} examines feature-noise robustness on the fixed split.

\subsubsection{Datasets and partitions}
Friedman~1 is a synthetic benchmark designed to explicitly assess a model's ability to capture nonlinear feature interactions~\cite{friedman1991multivariate}. It consists of $10$ independent input features, where the target variable is generated by known nonlinear interactions among a subset of features,
\begin{equation}
 y=10\sin(\pi x_1x_2)+20(x_3-0.5)^2+10x_4+5x_5+\epsilon.
 \label{eq:friedman}
\end{equation}
Inputs are uniform on $[0,1]^{10}$ and $\epsilon=0$ in all Friedman experiments; the last five features are irrelevant to the response.

Concrete Compressive Strength relates material composition to compressive strength~\cite{yeh1998modeling}. Its predictors describe mixture components and age.

California Housing is based on U.S.\ census data, featuring socio-economic and geographic factors~\cite{pace1997sparse}. The target is median house value, measured in units of \$100,000.

Energy Efficiency uses the heating-load target, $Y_1$, from building-design variables~\cite{tsanas2012accurate}. Yacht Hydrodynamics predicts residuary resistance per unit displacement from six hull and operating variables~\cite{yacht_hydrodynamics_243}. Auto MPG predicts fuel economy from vehicle attributes~\cite{auto_mpg_9}. We remove six Auto MPG rows with missing horsepower, exclude car names, and one-hot encode origin, leaving 392 rows and nine inputs.

Table~\ref{tab:datasets} gives the partition sizes for all tasks. Scientific benchmark partitions use nominal proportions 70/15/15 for Friedman, Energy, Yacht, and Auto MPG, 60/20/20 for Concrete, and 80/10/10 for California. For each of the five split seeds, all models receive identical partitions and independently fitted training scalers. Thus each scientific dataset/model combination comprises 25 fits. The synthetic recovery and context tasks are described in Section~\ref{sec:extended-experiments}.

\subsubsection{Predictive results}
CV-PURe achieves the lowest cross-split mean RMSE on Friedman, California, Energy, and Auto MPG (Table~\ref{tab:scientific}). Its Friedman RMSE is 0.0781, a 53.9\% reduction relative to MLP and 57.9\% relative to CV-MLP. It leads every individual split on Friedman and Energy, showing that these gains persist across training and test compositions. On Energy, its mean RMSE is 13.6\% lower than MLP's. California and Auto MPG extend the predictive benefit to socio-economic and vehicle-efficiency data.

ResMLP has the lowest cross-split mean on Concrete, and MLP on Yacht. The cross-split comparison exposes this task dependence while establishing a particularly strong advantage on the interaction-driven Friedman response. The following recovery experiments examine the learned structure behind that advantage.

\subsection{Ablation study: residual connections}
\label{sec:skip-ablation}

To examine the contribution of residual connections on Concrete, we compare each PURe model with a PU counterpart that removes both identity shortcuts. The branch transformations, readout, width, and training protocol are unchanged, and matched seeds start from identical parameter tensors. We evaluate clean inputs and Gaussian feature noise at $\sigma=0.05$ using the protocol in Section~\ref{sec:noise}.

Residual connections lower mean error and run-to-run variability in both real and complex networks (Table~\ref{tab:skip-ablation}). CV-PURe achieves the lowest MSE in both conditions. Relative to CV-PU, its paired MSE improvement is 0.0116 on clean inputs (95\% CI $[0.0024,0.0185]$) and 0.0110 under noise ($[0.0023,0.0177]$). The corresponding RV improvements are smaller, 0.0024 and 0.0019. These controlled comparisons establish the value of preserving the hidden state around the multiplicative transformations, with the clearest predictive gain in the complex network.

\begin{table}[pos=h]
\caption{Residual-connection ablation on Concrete: standardized-target test MSE (mean $\pm$ sample SD, five paired training seeds). PU removes only the identity shortcuts.}
\label{tab:skip-ablation}\centering\small
\setlength{\tabcolsep}{5pt}
\begin{tabular}{@{}lcrr@{}}
\toprule
Model & Shortcuts & $\sigma=0$ & $\sigma=0.05$ \\ \midrule
RV-PU & No & $0.1180\pm0.0081$ & $0.1367\pm0.0059$ \\
RV-PURe & Yes & $0.1157\pm0.0042$ & $0.1348\pm0.0042$ \\
CV-PU & No & $0.1226\pm0.0081$ & $0.1385\pm0.0071$ \\
CV-PURe & Yes & $\mathbf{0.1110\pm0.0048}$ & $\mathbf{0.1275\pm0.0044}$ \\
\bottomrule
\end{tabular}
\end{table}

\section{Interaction learning and interpretation}\label{sec:extended-experiments}
We connect prediction with learned interaction structure through repeated benchmark maps, exact recovery on known functions, and context-dependent applications.

\subsection{Interpretability and feature interaction}
\label{sec:scientific-interactions}
\begin{figure*}[pos=t]\centering
\includegraphics[width=\textwidth]{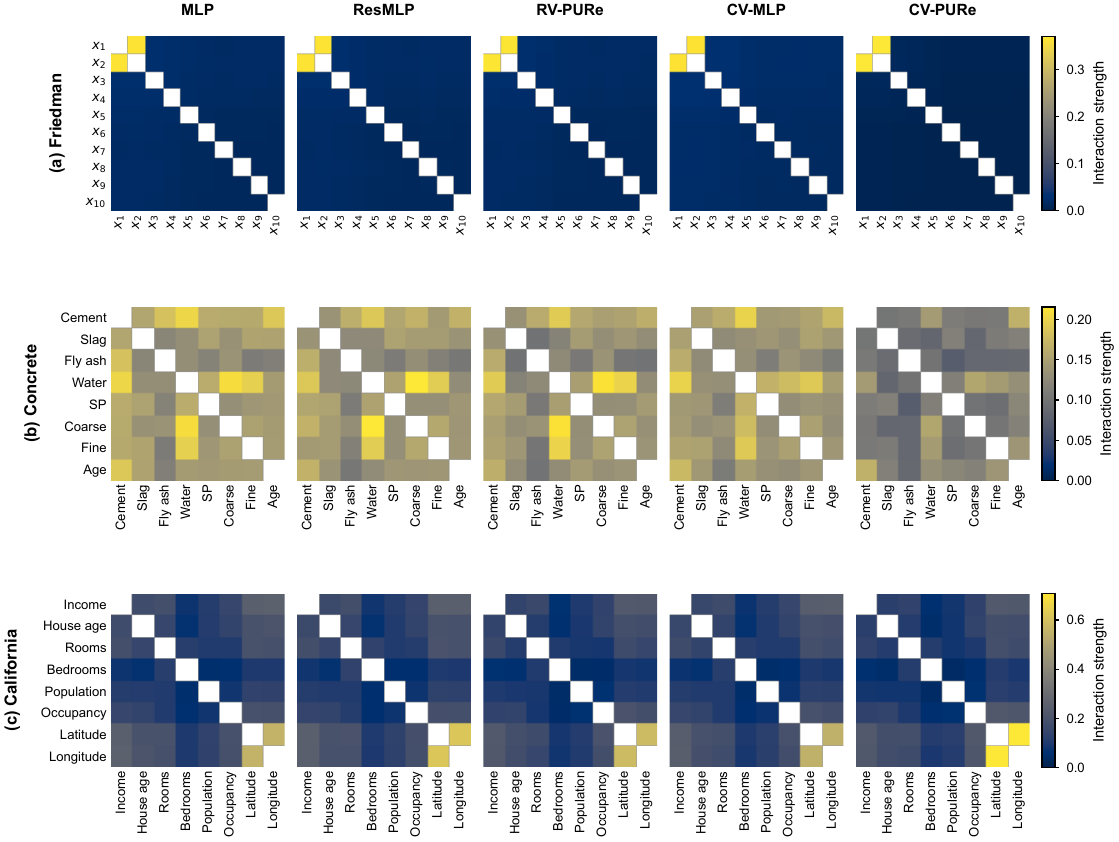}
\caption{Finite-difference interaction maps on Friedman 1 (top), Concrete Compressive Strength (middle), and California Housing (bottom). Off-diagonal cells show mean absolute mixed contrasts, averaged over five training seeds and five sampling repetitions. Color scales are shared across models within each dataset. SP denotes superplasticizer; Coarse/Fine denote aggregate components.}
\label{fig:scientific-interactions}\end{figure*}

To evaluate the interpretability and structural alignment of the models, we analyze pairwise feature interaction patterns using the finite-difference score. Figure~\ref{fig:scientific-interactions} shows the interaction maps for MLP, ResMLP, RV-PURe, CV-MLP, and CV-PURe on Friedman, Concrete, and California. We quantify their concentration and repeatability, and validate the strongest pair against the known Friedman response.

\begin{table*}[pos=t]
\caption{Interaction concentration, Friedman true-pair share, and Top-10 Jaccard stability. Except for mean between-run training stability, entries are mean $\pm$ sample SD over five training seeds after averaging five sampling repetitions. True-pair share uses the known Friedman coupling.}
\label{tab:scientific-interactions}\centering\small
\setlength{\tabcolsep}{10pt}
\begin{tabular}{@{}llrrrrr@{}}
\toprule
Dataset & Model & $H_{10}$ & $C_{10}$ (\%) & True pair (\%) & \shortstack{Jaccard\\Training} & \shortstack{Jaccard\\Sampling} \\ \midrule
Friedman & MLP & $0.646\pm0.020$ & $49.64\pm1.62$ & $30.95\pm1.87$ & $0.791$ & $0.699\pm0.069$ \\
 & ResMLP & $0.635\pm0.014$ & $49.60\pm1.47$ & $31.40\pm1.52$ & $0.674$ & $0.577\pm0.066$ \\
 & RV-PURe & $0.635\pm0.017$ & $49.80\pm1.54$ & $31.53\pm1.67$ & $0.724$ & $0.618\pm0.090$ \\
 & CV-MLP & $0.686\pm0.005$ & $46.63\pm0.67$ & $27.44\pm0.58$ & $0.674$ & $0.610\pm0.076$ \\
 & CV-PURe & $\mathbf{0.391\pm0.044}$ & $\mathbf{74.87\pm4.64}$ & $\mathbf{60.66\pm5.76}$ & $\mathbf{0.836}$ & $\mathbf{0.817\pm0.069}$ \\
\midrule
Concrete & MLP & $0.997\pm0.002$ & $44.18\pm1.19$ & --- & $0.464$ & $0.595\pm0.143$ \\
 & ResMLP & $0.996\pm0.003$ & $44.06\pm1.50$ & --- & $0.505$ & $\mathbf{0.622\pm0.108}$ \\
 & RV-PURe & $\mathbf{0.995\pm0.003}$ & $\mathbf{44.97\pm1.97}$ & --- & $0.391$ & $0.555\pm0.124$ \\
 & CV-MLP & $0.997\pm0.001$ & $42.97\pm0.83$ & --- & $0.498$ & $0.533\pm0.071$ \\
 & CV-PURe & $0.996\pm0.001$ & $44.90\pm0.86$ & --- & $\mathbf{0.581}$ & $0.604\pm0.026$ \\
\midrule
California & MLP & $0.966\pm0.003$ & $56.03\pm2.23$ & --- & $0.873$ & $0.818\pm0.115$ \\
 & ResMLP & $0.955\pm0.010$ & $56.36\pm1.18$ & --- & $0.858$ & $0.814\pm0.071$ \\
 & RV-PURe & $0.952\pm0.010$ & $58.27\pm2.08$ & --- & $0.858$ & $\mathbf{0.848\pm0.055}$ \\
 & CV-MLP & $0.960\pm0.010$ & $55.98\pm2.54$ & --- & $\mathbf{0.927}$ & $0.809\pm0.060$ \\
 & CV-PURe & $\mathbf{0.934\pm0.014}$ & $\mathbf{60.19\pm1.90}$ & --- & $0.806$ & $0.823\pm0.045$ \\
\bottomrule
\end{tabular}
\end{table*}

\paragraph{Sampling protocol.}
For each of five training runs on split 0, we draw five shared attribution samples using sampling seeds 0--4. Each sample pairs $M=\min(256,n_{\mathrm{test}})$ test rows drawn without replacement with training references drawn with replacement. We score all unordered input pairs.

\paragraph{Concentration and stability.}
Table~\ref{tab:scientific-interactions} reports normalized entropy $H_{10}$, Top-10 share $C_{10}$, true-pair share on Friedman, and Jaccard stability. For the ten strongest unordered pairs, $\mathcal{T}_{10}$, we normalize interaction strengths as
\begin{equation}
p_{ij}=\frac{S_{ij}}{\sum_{(a,b)\in\mathcal{T}_{10}}S_{ab}}.
\end{equation}

Normalized entropy and the Top-10 share are
\begin{align}
H_{10}&=-\frac{\sum_{(i,j)\in\mathcal{T}_{10}}p_{ij}\log p_{ij}}{\log 10},\nonumber\\
C_{10}&=\frac{\sum_{(i,j)\in\mathcal{T}_{10}}S_{ij}}{\sum_{i<j}S_{ij}}.
\label{eq:interaction-concentration}
\end{align}
Lower $H_{10}$ indicates greater concentration within the ten leading pairs, while $C_{10}$ measures their contribution to total strength. The Friedman true-pair share is $S_{12}/\sum_{i<j}S_{ij}$. Statistics are averaged over sampling seeds within checkpoints, then summarized across training seeds.

Stability uses Top-10 Jaccard overlap, $J(A,B)=|A\cap B|/|A\cup B|$. Training stability averages the ten overlaps among five sampling-averaged checkpoint maps. Sampling stability averages the ten overlaps among sampling repetitions within each checkpoint, followed by mean and sample SD over training seeds.

\paragraph{Interaction patterns and known-pair validation.}
On Friedman, the MLP baselines distribute interaction strength beyond the true $(x_1,x_2)$ coupling (Fig.~\ref{fig:scientific-interactions}, top).

Every model ranks the true pair first in all 25 training/sampling combinations. CV-PURe assigns 60.66\% of total strength to this pair, compared with 30.95\% for MLP and 27.44\% for CV-MLP. Its paired increase is 29.70 percentage points over MLP (95\% CI $[25.72,33.88]$) and 33.22 over CV-MLP ($[28.23,38.10]$). Absolute true-pair strength remains similar across models (0.363--0.370), so the stronger concentration reflects reduced strength assigned to pairs absent from the response. CV-PURe also has the highest training and sampling stability on Friedman. Together, these results show a more selective and repeatable representation of the known dependency.

Concrete maps emphasize water interactions with mixture components, and Top-10 entropy is near one. RV-PURe has the largest Top-10 share, CV-PURe the highest training stability, and ResMLP the highest sampling stability.

For the California Housing dataset (bottom panel), interaction structure is inherently more heterogeneous, and all models highlight the key spatial dependency between Latitude and Longitude.

CV-PURe has the largest Top-10 share, 60.19\%, while CV-MLP and RV-PURe lead training and sampling stability, respectively. These maps characterize how the predictors organize heterogeneous input dependencies alongside their predictive performance.

\subsection{Ground-truth interaction recovery}
\label{sec:interaction-recovery}

\subsubsection{Synthetic tasks and comparison models}
The response functions are
\begin{align}
 y_{\mathrm{pair}}&=x_1x_2+x_3x_4+\epsilon,\label{eq:pairwise}\\
 y_{\mathrm{higher}}&=x_1x_2x_3+x_4x_5x_6x_7+\epsilon.\label{eq:higher}
\end{align}
Inputs are independent and uniform on $[-1,1]$, with 20 distractors per task (24/27 features). Each task has 10,000 observations and Gaussian response noise with SD equal to 5\% of the generated clean signal's SD. Data generation and the 70/15/15 split are fixed across models. Models use width 64, training seeds 0--2, data-generation seed 0, split seed 0, and the shared training protocol. The pairwise task compares MLP, ResMLP, RV-PURe, and CV-PURe; the higher-order task adds CV-MLP with the same width and real-valued projection as CV-PURe.

The pairwise ground truth is $\mathcal{T}_{\mathrm{pair}}=\{(1,2),(3,4)\}$. For the higher-order task, $\mathcal{T}_{\mathrm{higher}}$ comprises the nine pairwise projections within $\{1,2,3\}$ and $\{4,5,6,7\}$.

\subsubsection{Interaction scoring and recovery criteria}

We rank all pairs by $S_{ij}$ using five attribution samples of 256 test/reference pairs per checkpoint, drawn as in Section~\ref{sec:scientific-interactions}. Three training runs yield 15 rankings; prediction metrics summarize the three runs.

Let $\widehat{\mathcal{T}}_K$ be the first $K$ pairs in the ranking. We report
\begin{align}
 \mathrm{Precision@}K&=|\widehat{\mathcal{T}}_K\cap\mathcal{T}|/K,\nonumber\\
 \mathrm{Recall@}K&=|\widehat{\mathcal{T}}_K\cap\mathcal{T}|/|\mathcal{T}|.
\end{align}
With $K=|\mathcal{T}|$, precision and recall coincide. Mean rank averages the one-based positions of the true pairs, with ideal values 1.5 (two pairs) and 5 (nine pairs). Exact recovery means that the top $K$ contains all true pairs and no distractors.

\subsubsection{Predictive accuracy and recovered interactions}

All four models recover both true pairs in every repetition. CV-PURe gives the lowest RMSE, 0.0386 versus 0.0442 for MLP (Table~\ref{tab:recovery}). Thus predictive accuracy distinguishes the models even when all recover this pairwise structure exactly.

\begin{table*}[pos=t]
\caption{Synthetic prediction and interaction recovery. Mean $\pm$ sample SD uses three training seeds for RMSE and 15 training/sampling combinations for ranks. $K=2$ (pairwise) or $K=9$ (higher-order); Exact counts complete recoveries.}
\label{tab:recovery}\centering\small
\setlength{\tabcolsep}{10pt}
\begin{tabular}{@{}llrrrr@{}}
\toprule
Task & Model & Test RMSE & Precision@$K$ = Recall@$K$ & Mean rank & Exact \\ \midrule
Pairwise & MLP & $0.0442\pm0.0005$ & $\mathbf{1.000}$ & $\mathbf{1.5\pm0.0}$ & \textbf{15/15} \\
 & ResMLP & $0.0440\pm0.0018$ & $\mathbf{1.000}$ & $\mathbf{1.5\pm0.0}$ & \textbf{15/15} \\
 & RV-PURe & $0.0402\pm0.0017$ & $\mathbf{1.000}$ & $\mathbf{1.5\pm0.0}$ & \textbf{15/15} \\
 & CV-PURe & $\mathbf{0.0386\pm0.0021}$ & $\mathbf{1.000}$ & $\mathbf{1.5\pm0.0}$ & \textbf{15/15} \\
\midrule
Higher-order & MLP & $0.1351\pm0.0028$ & $0.333$ & $93.5\pm33.2$ & 0/15 \\
 & ResMLP & $0.1338\pm0.0047$ & $0.333$ & $75.9\pm10.5$ & 0/15 \\
 & RV-PURe & $0.1315\pm0.0037$ & $0.333$ & $93.0\pm30.8$ & 0/15 \\
 & CV-MLP & $0.1376\pm0.0023$ & $0.333$ & $121.3\pm23.2$ & 0/15 \\
 & CV-PURe & $\mathbf{0.0797\pm0.0072}$ & $\mathbf{1.000}$ & $\mathbf{5.0\pm0.0}$ & \textbf{15/15} \\
\bottomrule
\end{tabular}
\end{table*}

On the higher-order task, CV-PURe achieves the lowest RMSE and recovers all nine pairwise projections in every repetition; other models recover one third in their top-nine rankings. Its RMSE improvement over CV-MLP is 0.0579 (95\% CI \([0.0491,0.0664]\)), a 42.1\% reduction. CV-MLP uses the same width and signed readout, with only 256 more real scalar parameters. These joint gains support multiplicative feature construction as an effective inductive bias for this task.

\subsection{Context-dependent prediction and interpretation}
\label{sec:context-case-studies}
Bike Sharing and Parkinsons examine how measurements combine with temporal and subject context.

\subsubsection{Inputs, comparison models, and evaluation}
\label{sec:context-models}

Inputs comprise ordinary measurements $\mathbf{u}$ and context $\mathbf{c}$. Both PURe variants and CV-MLP receive the encoded, scaled vector $\mathbf{x}=[\mathbf{u};\mathbf{c}]$ and follow Section~\ref{sec:shared-protocol}, with complex models embedding inputs as $\mathbf{x}+\mathrm{i}\mathbf{0}$.

MLP-ordinary receives only $\mathbf{u}$, whereas MLP-context receives $\mathbf{x}=[\mathbf{u};\mathbf{c}]$. MLP-explicit-cross augments this input with all ordinary--context products:
\begin{equation}
 \mathbf{x}_{\mathrm{cross}}
 =
 [\mathbf{u};\mathbf{c};
 \operatorname{vec}(\mathbf{u}\mathbf{c}^{\mathsf T})].
 \label{eq:explicit-cross}
\end{equation}
These products are formed after input standardization, without selecting individual crosses.

DCNv2-context combines three full-rank cross layers with a parallel three-layer ReLU network of width 64~\cite{wang2021dcn}. Starting from $\mathbf{x}_0=\mathbf{x}$, the cross branch computes
\begin{equation}
 \mathbf{x}_{\ell+1}
 =
 \mathbf{x}_0 \odot
 (W_{\ell}\mathbf{x}_{\ell}+\mathbf{b}_{\ell})
 +\mathbf{x}_{\ell},
 \label{eq:dcnv2-cross}
\end{equation}
where $\ell=0,1,2$, $\odot$ denotes element-wise multiplication, and $W_{\ell}\in\R^{p\times p}$. The $p$-dimensional cross output and 64-dimensional deep output are concatenated before a scalar affine readout.

All MLP and PURe hidden widths are 64; DCNv2 retains dimension $p$ in its cross branch. The ordinary, full-context, and explicit-cross inputs have dimensions 4, 26, and 114 on Bike Sharing, and 16, 61, and 781 on Parkinsons. Table~\ref{tab:context} reports the Bike Sharing predictive comparison over five training seeds.

\begin{table*}[pos=tp]
\caption{Bike Sharing prediction (mean $\pm$ sample SD, five training seeds). MAE and RMSE are in rental counts.}
\label{tab:context}\centering\small
\setlength{\tabcolsep}{10pt}
\begin{tabular}{@{}lrrr@{}}
\toprule
Model & MAE & RMSE & $R^2$ \\ \midrule
MLP-ordinary & $145.6385\pm1.2411$ & $207.3371\pm2.2520$ & $0.1156\pm0.0191$ \\
MLP-context & $69.5852\pm4.5626$ & $99.8035\pm7.2218$ & $0.7942\pm0.0300$ \\
MLP-explicit-cross & $77.5185\pm4.1527$ & $115.6913\pm6.4432$ & $0.7240\pm0.0311$ \\
DCNv2-context & $82.5452\pm20.8131$ & $109.7941\pm25.3903$ & $0.7414\pm0.1196$ \\
RV-PURe & $64.3894\pm4.9263$ & $92.3034\pm9.1224$ & $0.8234\pm0.0353$ \\
CV-MLP & $67.8937\pm9.3982$ & $98.2697\pm12.3125$ & $0.7989\pm0.0506$ \\
CV-PURe & $\mathbf{57.8725\pm7.1848}$ & $\mathbf{82.4341\pm6.9617}$ & $\mathbf{0.8594\pm0.0236}$ \\
\bottomrule
\end{tabular}
\end{table*}

\begin{figure*}[pos=tp]\centering
\includegraphics[width=0.95\textwidth]{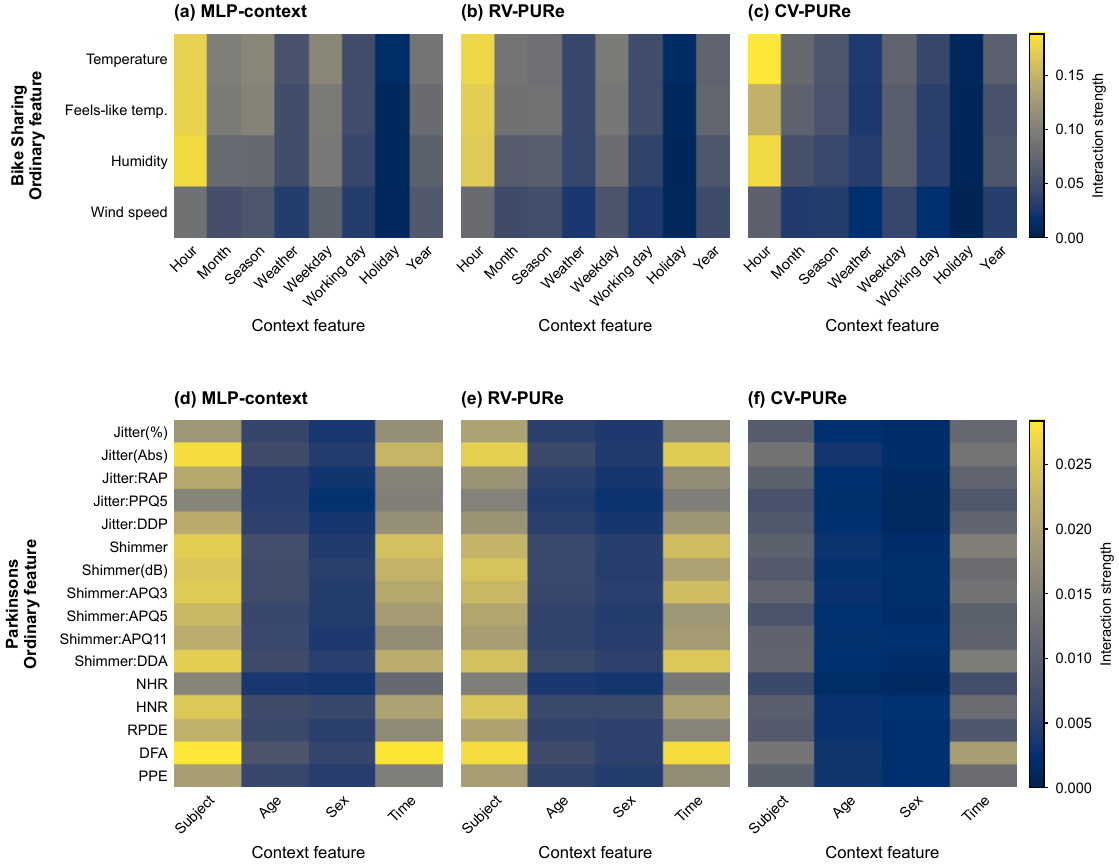}
\caption{Feature-context interaction maps for Bike Sharing (top) and Parkinsons (bottom). Grouped finite-difference scores average five training runs, with shared samples and within-dataset color scales.}
\label{fig:context}
\end{figure*}

\paragraph{Interaction maps.}
\label{sec:context-interpretability}
We jointly replace each variable's encoded columns in Eqs.~\eqref{eq:finite-difference}--\eqref{eq:interaction-score}. Figure~\ref{fig:context} compares MLP-context, RV-PURe, and CV-PURe, averaging training seeds 0--4 using one shared sample of 256 test/reference pairs with sampling seed 0, drawn as in Section~\ref{sec:scientific-interactions}.

\subsubsection{Bike Sharing}\label{sec:bike-sharing}
\paragraph{Data and inputs.}
The hourly Bike Sharing data contain 17,379 observations of rental counts~\cite{fanaee2014event,bike_sharing_275}. We order observations by time and use the first 60\% for training, the next 20\% for validation, and the final 20\% for testing. Ordinary inputs are temperature, apparent temperature, humidity, and wind speed. Context comprises cyclic hour/month encodings, one-hot season, weather category, and weekday, plus working-day, holiday, and year indicators. The row index, raw date, and casual/registered counts are excluded; the latter two sum to the target.

\paragraph{Predictive results.}
Adding context approximately halves MLP mean MAE, from 145.64 to 69.59 rentals (Table~\ref{tab:context}). CV-PURe achieves the best mean MAE, RMSE, and $R^2$. Its RMSE is 82.4341, improving on MLP-context by 17.3694 rentals (95\% CI $[7.7713,25.4227]$) and on CV-MLP by 15.8356 ($[5.1553,27.1371]$). These correspond to reductions of 17.4\% and 16.1\%, respectively. The MAE improvements are 11.7127 over MLP-context ($[4.4735,18.9519]$) and 10.0212 over CV-MLP ($[0.4455,19.6547]$). Appending all ordinary--context products or using DCNv2 does not improve on MLP-context here; CV-PURe obtains a stronger predictive representation from the same full-context inputs.

\paragraph{Context contributions and interaction structure.}
Table~\ref{tab:ablations} summarizes context ablations for RV-PURe and CV-PURe on Bike Sharing. We retrain each model after removing one of three context groups, using training seeds 0--2 and pairing each ablation with its full-context fit. Year and four ordinary weather measurements remain available. Calendar comprises weekday, holiday, and working day.

\begin{table*}[pos=t]
\caption{RV-PURe and CV-PURe context ablations on Bike Sharing (mean $\pm$ sample SD, training seeds 0--2). $\Delta$MAE is ablated minus full-context MAE, paired by seed. Calendar comprises weekday, holiday, and working day.}
\label{tab:ablations}\centering\small
\setlength{\tabcolsep}{10pt}
\begin{tabular}{@{}llrr@{}}
\toprule
Removed group & Model & MAE & $\Delta$MAE [95\% CI] \\ \midrule
None & RV-PURe & $65.4150\pm5.1311$ & --- \\
None & CV-PURe & $\mathbf{57.0821\pm4.8786}$ & --- \\
\midrule
Hour, month & RV-PURe & $143.4297\pm3.0181$ & $78.0148\ [74.7200,80.5912]$ \\
Hour, month & CV-PURe & $\mathbf{140.6423\pm1.6397}$ & $83.5602\ [77.4822,88.1939]$ \\
\midrule
Calendar & RV-PURe & $93.6811\pm12.8377$ & $28.2661\ [13.0414,44.5532]$ \\
Calendar & CV-PURe & $\mathbf{82.1527\pm2.8624}$ & $25.0706\ [19.6722,33.2718]$ \\
\midrule
Season, weather category & RV-PURe & $\mathbf{65.8865\pm4.3071}$ & $0.4716\ [-0.4980,1.0013]$ \\
Season, weather category & CV-PURe & $65.9548\pm6.2430$ & $8.8726\ [4.9890,11.3940]$ \\
\bottomrule
\end{tabular}
\end{table*}

Removing hour/month produces the largest error increase for both models. CV-PURe's MAE rises from 57.0821 to 140.6423, a paired increase of 83.5602 rentals. Removing calendar indicators raises its MAE by 25.0706. Removing season/weather categories raises CV-PURe's MAE by 8.8726 (95\% CI \([4.9890,11.3940]\)), versus 0.4716 for RV-PURe (\([-0.4980,1.0013]\)). All three CV-PURe ablations increase error for every training seed, indicating predictive contributions from each context group, especially temporal cycles.

Figure~\ref{fig:context}(a--c) shows strong hour interactions with temperature, apparent temperature, and humidity across all three models; wind-speed and holiday interactions are weaker. These maps reveal non-additive weather--hour dependencies, while paired ablations establish the predictive value of the context inputs. The seasonal ablation further shows that categorical context contributes beyond the retained numerical weather measurements.

\subsubsection{Parkinsons Telemonitoring}\label{sec:parkinsons}
\paragraph{Data and inputs.}
Parkinsons Telemonitoring contains 5,875 voice recordings from 42 subjects~\cite{tsanas2009accurate,parkinsons_telemonitoring_189}. We predict the supplied total Unified Parkinson's Disease Rating Scale (UPDRS) score, which is linearly interpolated between clinical assessments in the released dataset. The 16 voice measurements form the ordinary inputs. Context includes subject identity (42 one-hot columns), age, sex, and elapsed recording time. Motor UPDRS is excluded because it is another clinical target.

Recordings are split chronologically within each subject using nominal 60/20/20 boundaries. Records with the same subject and timestamp stay together, giving the actual partition sizes in Table~\ref{tab:datasets}. The task predicts later recordings of the same 42 subjects.

\paragraph{Predictive results.}
Context reduces MLP mean RMSE from 9.6313 to 4.5245. CV-PURe reaches $4.1841\pm0.1128$, a paired improvement of 0.3404 UPDRS points over MLP-context (95\% CI $[0.2285,0.4506]$). CV-MLP reaches a similar RMSE of $4.1607\pm0.1801$ and has the lower MAE. The learning-curve analysis in Section~\ref{sec:sample-efficiency} examines how these models use reduced training data.

\paragraph{Interaction interpretation.}
In Fig.~\ref{fig:context}(d--f), MLP-context and RV-PURe show strong subject/time interactions with voice measurements; age and sex interactions are weaker. CV-PURe assigns lower strength to subject--voice pairs while retaining prominent elapsed-time interactions, particularly for detrended fluctuation analysis and shimmer measurements. The context maps therefore capture distinct ways of combining longitudinal measurements with subject information.

\section{Robustness, sample efficiency, and computational cost}
\label{sec:practical-evaluation}

\subsection{Robustness under feature noise}\label{sec:noise}
\label{sec:robustness-sample-efficiency}
To assess robustness, predictive performance is evaluated under both clean conditions and Gaussian noise corruption, where zero-mean noise with standard deviation $\sigma\in\{0.02,0.05,0.10\}$ is added to selected ordinary numerical input features at test time. This tests measurement-noise sensitivity across corruption severities~\cite{hendrycks2019benchmarking}.

We evaluate five scientific and seven context models over five training runs without retraining, including the clean endpoint (\(\sigma=0\)). Noise is added in standardized space to ordinary numerical columns with more than two distinct training values, including model year; context, binary, and one-hot columns remain fixed. Explicit crosses are recomputed after corruption. Five corruption realizations, with seeds 0--4, are shared across models at each nonzero noise level; clean inputs are evaluated once. All scientific noise curves use split seed 0.

To combine datasets with different target units, let $s_{y,d}$ be the
population standard deviation of the test targets on dataset $d$ and
define
$e_{d,m,s}(\sigma)=\mathrm{RMSE}_{d,m,s}(\sigma)/s_{y,d}$
for model $m$ and training seed $s$, after averaging RMSE across
corruption realizations.

\begin{figure*}[pos=tp]\centering
\includegraphics[width=\textwidth]{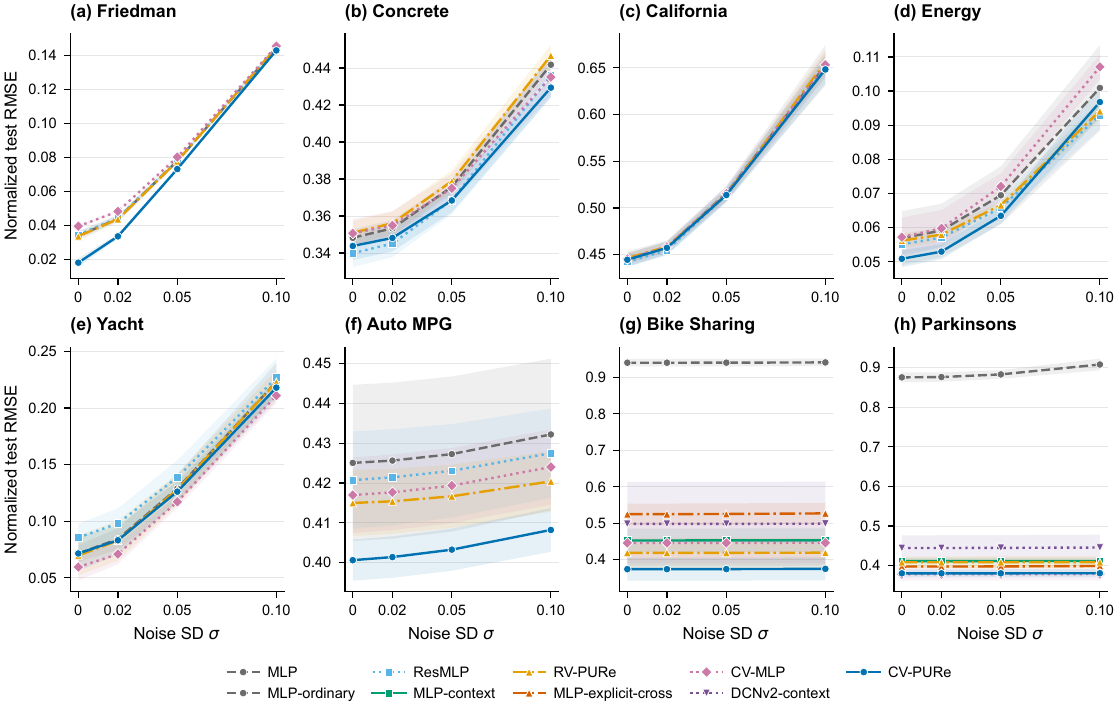}
\caption{Feature-noise curves over $\sigma=0$--$0.10$: test RMSE normalized by test-target SD. Lines and bands show mean and sample SD across five training seeds, after averaging five corruption realizations per nonzero level.}
\label{fig:noise}\end{figure*}

Error-curve AUC has been used to summarize performance across label-corruption levels~\cite{hendrycks2019using}. For feature noise, we integrate normalized RMSE over $\sigma_0=0$, $\sigma_1=0.02$, $\sigma_2=0.05$, and $\sigma_3=0.10$, dividing by the interval length:
\begin{equation}
\begin{aligned}
A_{d,m,s}={}&\frac{1}{0.10}
\sum_{k=0}^{2}
\frac{e_{d,m,s}(\sigma_k)+e_{d,m,s}(\sigma_{k+1})}{2}
\\
&\qquad\times(\sigma_{k+1}-\sigma_k).
\end{aligned}
\label{eq:noise-auc}
\end{equation}
We average $A_{d,m,s}$ equally across datasets within each suite, then summarize the five training-seed values. Table~\ref{tab:noise} reports the interval summaries for the five scientific and six context-input models; Fig.~\ref{fig:noise} shows dataset-level curves and also includes MLP-ordinary.

\paragraph{Results.}
CV-PURe has the lowest mean AUC in both suites (Table~\ref{tab:noise}). Its paired scientific-suite improvements are 0.0081 over MLP (95\% CI $[0.0038,0.0122]$) and 0.0054 over CV-MLP ($[0.0038,0.0071]$). Context-suite improvements are 0.0548 over MLP-context ($[0.0365,0.0731]$) and 0.0349 over CV-MLP ($[0.0050,0.0671]$). These summaries show that its predictive advantage extends across the tested input-noise range.

\begin{table}[pos=h]
\caption{Noise-performance AUC over $\sigma=0$--$0.10$, normalized by test-target SD and averaged equally across datasets in each suite. Mean $\pm$ sample SD and bootstrap intervals summarize five training seeds. Lower is better.}
\label{tab:noise}\centering\small
\setlength{\tabcolsep}{3pt}
\begin{tabular}{@{}lrr@{}}
\toprule
Model & AUC & 95\% CI \\ \midrule
\multicolumn{3}{@{}l}{Scientific suite (six datasets)} \\
MLP & $0.2718\pm0.0050$ & $[0.2676,0.2752]$ \\
ResMLP & $0.2704\pm0.0033$ & $[0.2677,0.2730]$ \\
RV-PURe & $0.2700\pm0.0035$ & $[0.2673,0.2726]$ \\
CV-MLP & $0.2692\pm0.0022$ & $[0.2677,0.2709]$ \\
CV-PURe & $\mathbf{0.2637\pm0.0019}$ & $[0.2623,0.2652]$ \\
\midrule
\multicolumn{3}{@{}l}{Context suite (two datasets)} \\
MLP-context & $0.4321\pm0.0171$ & $[0.4205,0.4467]$ \\
MLP-explicit-cross & $0.4618\pm0.0142$ & $[0.4506,0.4725]$ \\
DCNv2-context & $0.4715\pm0.0605$ & $[0.4298,0.5194]$ \\
RV-PURe & $0.4136\pm0.0180$ & $[0.3980,0.4281]$ \\
CV-MLP & $0.4121\pm0.0297$ & $[0.3900,0.4359]$ \\
CV-PURe & $\mathbf{0.3773\pm0.0177}$ & $[0.3639,0.3903]$ \\
\bottomrule
\end{tabular}
\end{table}

CV-PURe retains the lowest mean error at every noise level on Friedman, Auto MPG, and Bike Sharing (Fig.~\ref{fig:noise}). Friedman curves approach one another as noise increases. Energy favors CV-PURe at lower noise levels and ResMLP at $\sigma=0.10$; Concrete and California favor CV-PURe at the higher noise levels. CV-MLP leads on Yacht and Parkinsons. The context curves are comparatively flat when ordinary measurements are perturbed and context is held fixed.

\subsection{Sample efficiency on three datasets}\label{sec:sample-efficiency}
\begin{figure*}[pos=tp]\centering
\includegraphics[width=0.95\textwidth]{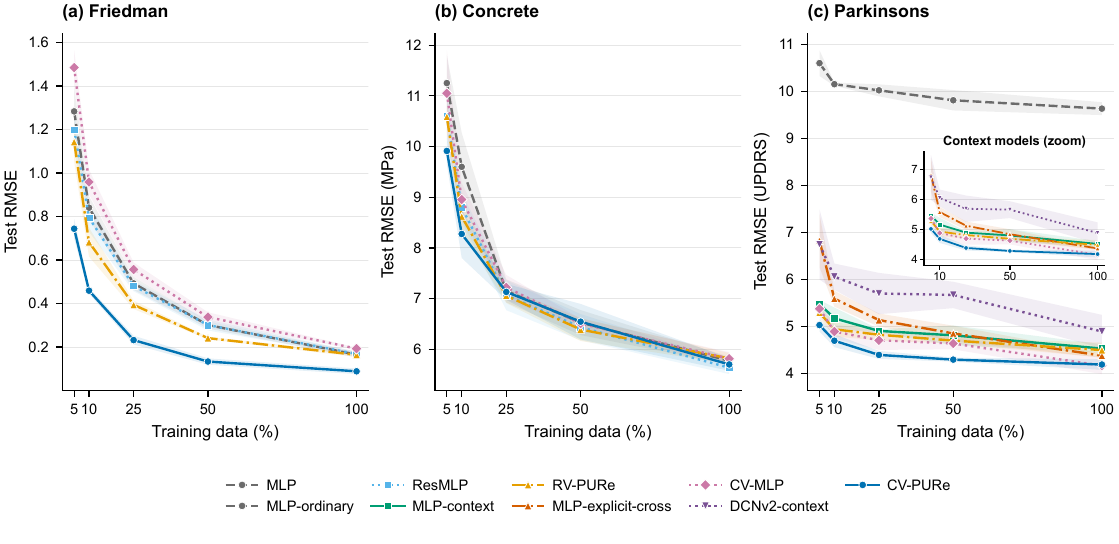}
\caption{Sample efficiency on Friedman, Concrete, and Parkinsons. Test RMSE is shown as mean \(\pm\) sample SD over five training seeds at nested training fractions of 5\%--100\%. Inset: Parkinsons context models.}
\label{fig:sample}\end{figure*}

To evaluate sample efficiency, we train the comparison models on progressively smaller subsets of Friedman, Concrete, and Parkinsons while keeping the validation and test sets fixed. Figure~\ref{fig:sample} shows nested learning curves~\cite{mohr2024learning}; Table~\ref{tab:sample} summarizes their RMSE area under the learning curve (AULC).

\begin{table}[pos=h]
\caption{RMSE area under the learning curve over training fractions 0.05--1.00 (mean $\pm$ sample SD, five training seeds). Lower is better.}
\label{tab:sample}\centering\small
\setlength{\tabcolsep}{3pt}
\begin{tabular}{@{}llr@{}}
\toprule
Dataset & Model & AULC \\ \midrule
Friedman & MLP & $0.3692\pm0.0164$ \\
 & ResMLP & $0.3608\pm0.0079$ \\
 & RV-PURe & $0.3073\pm0.0123$ \\
 & CV-MLP & $0.4195\pm0.0159$ \\
 & CV-PURe & $\mathbf{0.1833\pm0.0110}$ \\
\midrule
Concrete & MLP & $6.5632\pm0.1440$ \\
 & ResMLP & $\mathbf{6.3814\pm0.1678}$ \\
 & RV-PURe & $6.3898\pm0.1269$ \\
 & CV-MLP & $6.5106\pm0.1201$ \\
 & CV-PURe & $6.3816\pm0.1658$ \\
\midrule
Parkinsons & MLP-ordinary & $9.3707\pm0.1196$ \\
 & MLP-context & $4.5667\pm0.1029$ \\
 & MLP-explicit-cross & $4.6657\pm0.0815$ \\
 & DCNv2-context & $5.2593\pm0.3146$ \\
 & RV-PURe & $4.4724\pm0.0718$ \\
 & CV-MLP & $4.3401\pm0.1214$ \\
 & CV-PURe & $\mathbf{4.1267\pm0.0426}$ \\
\bottomrule
\end{tabular}
\end{table}

\paragraph{Training subsets and summary metric.}
Training subsets of 5\%, 10\%, 25\%, 50\%, and 100\% are nested using selection seed 0 and shared across models. Parkinsons samples are selected within subjects, rounding counts up to retain all 42 subjects. The nominal 5\% subsets contain 350, 31, and 202 observations for Friedman, Concrete, and Parkinsons, respectively. Input and target scalers are refitted on each subset; validation/test partitions and full-task optimization settings remain fixed. Each fraction uses training seeds 0--4, and the 100\% endpoints reuse the corresponding full-data fits. Comparisons include all five scientific models on Friedman/Concrete and all seven context models on Parkinsons.

Let $q_0,\ldots,q_4$ denote these training fractions. RMSE AULC is
\begin{equation}
L_{d,m,s}=\sum_{k=0}^{3}
\frac{r_{d,m,s}(q_k)+r_{d,m,s}(q_{k+1})}{2}
(q_{k+1}-q_k),
\label{eq:aulc}
\end{equation}
where $r_{d,m,s}(q)$ is test RMSE in original units for model $m$, dataset $d$, training seed $s$, and fraction $q$. AULC is not divided by the interval length and is compared within each dataset; lower is better.

\paragraph{Results.}
CV-PURe achieves the lowest mean AULC on Friedman and Parkinsons, and is nearly tied with ResMLP on Concrete (6.3816 versus 6.3814). Relative to CV-MLP, its paired AULC improvements are 0.2362 on Friedman (95\% CI $[0.2243,0.2482]$), 0.1290 on Concrete ($[0.0307,0.2218]$), and 0.2133 on Parkinsons ($[0.1197,0.3204]$). The Friedman AULC is 56.3\% lower than CV-MLP's, indicating a strong advantage across the range of available supervision.

CV-PURe has the lowest mean error at every training fraction on Friedman and at 5\%--50\% on Parkinsons. On Concrete it leads at 5\% and 10\%, RV-PURe leads at 25\% and 50\%, and ResMLP at 100\%. With only 31 Concrete training observations, CV-PURe reaches RMSE $9.9128\pm0.2654$ MPa versus $11.2521\pm0.5469$ for MLP. At 5\% on Parkinsons, it reaches $5.0237\pm0.1365$ versus $5.4736\pm0.0845$ for MLP-context. These results connect multiplicative representation learning to effective prediction with limited data.

\subsection{Computational efficiency}\label{sec:efficiency}
We report parameter counts and inference latency separately, as they describe different aspects of efficiency~\cite{dehghani2021efficiency}.

\paragraph{Model capacity.}
Table~\ref{tab:efficiency} combines capacity and latency. Each complex parameter counts as two real scalars, and the projection adds three real trainable parameters. CV-PURe has approximately half the parameters of the width-128 real scientific models. CV-MLP and CV-PURe differ by only 256 real scalar parameters, approximately 0.5\%, on every task with a matched input dimension and width.

\begin{table*}[pos=tp]
\caption{Model capacity and full-test inference latency on an RTX 5080. Complex parameters count as two real scalars; complex-model totals include three real projection parameters. Scientific parameter ranges reflect input dimensions. Latency is in milliseconds (mean $\pm$ sample SD of five checkpoint-level medians).}
\label{tab:efficiency}\centering\small
\setlength{\tabcolsep}{3pt}
\begin{tabular}{@{}llrrrr@{}}
\toprule
Task & Model & Real parameters & Friedman (1,501) & Yacht (47) & Auto MPG (60) \\ \midrule
Scientific & MLP & 100,097--100,609 & $4.8034\pm0.6594$ & $\mathbf{0.3810\pm0.0136}$ & $\mathbf{0.4344\pm0.0946}$ \\
 & ResMLP & 100,097--100,609 & $\mathbf{4.6084\pm0.1076}$ & $0.5335\pm0.1207$ & $0.4669\pm0.0792$ \\
 & RV-PURe & 99,841--100,353 & $5.6412\pm0.4725$ & $0.5010\pm0.1009$ & $0.6007\pm0.1121$ \\
 & CV-MLP & 50,949--51,461 & $12.4122\pm1.8542$ & $0.9592\pm0.0232$ & $0.9533\pm0.0134$ \\
 & CV-PURe & 50,693--51,205 & $10.4210\pm1.7786$ & $0.7656\pm0.0176$ & $1.0488\pm0.0477$ \\
\midrule
Higher-order & CV-MLP & 53,637 & --- & --- & --- \\
 & CV-PURe & 53,381 & --- & --- & --- \\
\midrule
Bike Sharing & CV-MLP & 53,509 & --- & --- & --- \\
 & CV-PURe & 53,253 & --- & --- & --- \\
\midrule
Parkinsons & CV-MLP & 57,989 & --- & --- & --- \\
 & CV-PURe & 57,733 & --- & --- & --- \\
\bottomrule
\end{tabular}
\end{table*}

\paragraph{Inference protocol.}
Full-test inference uses one NVIDIA GeForce RTX 5080, PyTorch 2.14.0, CUDA 13.0, and batch size 128. After at least 20 warm-up passes and 0.5 seconds, we time 100 passes with CUDA synchronization before and after each. Table~\ref{tab:efficiency} summarizes checkpoint-level medians across training seeds 0--4. Timing includes batching, CPU--GPU transfers, and evaluation, excluding checkpoint loading, preprocessing, and metric calculation.

\paragraph{Results.}
CV-PURe takes 10.4210 ms for the full Friedman test set, 0.7656 ms for Yacht, and 1.0488 ms for Auto MPG. These are approximately 2.0--2.4 times the corresponding MLP latencies. Complex arithmetic and logarithmic/exponential operations add work despite the smaller parameter count. CV-PURe is faster than CV-MLP on Friedman and Yacht, and slower on Auto MPG. The resulting tradeoff is a compact parameterization with stronger interaction-driven prediction and a higher evaluation cost than the real MLP.

\section{Analysis and discussion}\label{sec:discussion}

\subsection{Interaction-aligned inductive bias}
The central empirical finding is that product-unit models are beneficial when the target is governed by nonlinear feature interactions. Product units directly parameterize multiplicative couplings, so the model does not need to synthesize interaction structure indirectly through many additive compositions.

The benchmark and recovery experiments establish this connection at complementary levels. CV-PURe leads all five Friedman splits and reduces cross-split mean RMSE by more than half relative to MLP. Its interaction maps preserve the true coupling while reducing off-target strength, and their leading pairs are more repeatable across training and sampling. On the higher-order task, the same architecture combines a 42.1\% RMSE reduction over CV-MLP with recovery of all nine true pairwise projections in every repetition. This agreement between prediction and known structure is the central evidence for the proposed interaction bias.

The cross-split benchmark also identifies where other representations excel: ResMLP leads on Concrete and MLP on Yacht. Across the full comparison, CV-PURe leads four of six datasets, with especially strong gains on analytically defined interactions and building-energy prediction. Together with the known-function recovery results, these comparisons support product-unit residual networks as an effective architecture for learning nonlinear interactions.

\subsection{Complex-valued modeling}
Complex-valued product units often improve mean performance, potentially because magnitude and phase offer a richer parameterization for representing nonlinear dependencies. Equation~\eqref{eq:amplitude-phase} makes this coupling explicit: the learned exponents transform log-amplitude and phase jointly. The real-valued projection then maps the resulting complex representation onto the signed target scale. CV-MLP shares this readout and nearly the same parameter count, making the higher-order comparison particularly informative: additive complex layers recover three of the nine true pairs in their top-nine rankings, whereas CV-PURe recovers all nine.

\subsection{Why residual connections matter here}
The residual path carries the hidden state around each multiplicative transformation. The combination of both components in PURe yields the best overall results in the paired Concrete ablation (Table~\ref{tab:skip-ablation}). Removing only the identity shortcuts increases CV-PURe's mean error under both clean and noisy inputs. Both PURe variants also have lower run-to-run SD than their no-shortcut counterparts. Independently initialized product-unit blocks allow both branches to adapt during training, while the identity paths retain the accumulated representation.

\subsection{Learning to use context}
Bike Sharing provides a direct application of interaction learning. The full-context models receive the same temporal and categorical information; CV-PURe improves on MLP-context and CV-MLP under this shared input representation. The new CV-PURe ablations identify contributions from all three context groups, with hour/month producing the largest error increase and season/weather categories contributing beyond the retained numerical weather measurements. Weather--hour patterns shared across the three model maps connect these predictive contributions to non-additive input relationships. The comparison with explicit crosses and DCNv2 further shows the benefit of learning multiplicative latent factors for this task.

Parkinsons extends the analysis to longitudinal measurements. Subject and time information substantially improve prediction of later recordings from the same individuals. CV-PURe organizes voice--context dependencies differently from the real models and attains the lowest learning-curve AULC, with its advantage most visible when training recordings are scarce. Together, the two applications show how multiplicative representations combine measurements with the conditions under which they were collected.

\subsection{Noise, sample efficiency, and computational cost}
CV-PURe has the lowest mean noise-performance AUC in both suites, demonstrating strong predictive accuracy across the tested corruption levels. Dataset-level curves show how this advantage evolves with noise: it persists throughout the range on Friedman, Auto MPG, and Bike, while model ordering changes on Energy, Concrete, and California. The learning curves extend the evidence to reduced supervision, with clear AULC gains over CV-MLP on all three datasets and the largest reduction on Friedman. These benefits accompany a higher inference cost than the real MLP, despite CV-PURe's smaller scientific-model parameter count.

\subsection{Architectural and experimental development from the conference study}
The conference study~\cite{li2026modeling} used the magnitude readout \(\widehat{y}=|o|\), restricting network-scale predictions to nonnegative values. To support signed responses and centered, standardized targets, we adopt the trainable projection \(\widehat{y}=\alpha\Re(o)+\beta\Im(o)+b\) for both CV-PURe and CV-MLP throughout the present experiments. Both PURe residual blocks now use independent default initialization, while the final output affine map remains zero-initialized.

To improve reproducibility, all models were rerun in the same computational environment with explicitly specified seeds for data generation, splitting, training, and evaluation. The scientific benchmark now averages five data partitions, each evaluated over five training seeds. The updated numerical results therefore reflect changes to the readout, initialization, and evaluation protocol. These developments preserve the conference paper's product-unit residual foundation while strengthening the evidence through cross-split prediction, interaction recovery, and paired context analysis.

\section{Conclusion}\label{sec:conclusion}
We evaluated PURe, a product-unit residual architecture for tabular regression, and investigated both real- and complex-valued parameterizations through prediction and input-level interaction analysis. Across interaction-driven synthetic data and real-world benchmarks, product-unit models generally achieved lower prediction error than MLP baselines. With a signed real-valued projection and an explicit seed protocol, CV-PURe achieves the lowest cross-split mean RMSE on four of six scientific datasets.

The strongest contribution is the connection between predictive gains and recovered interaction structure. CV-PURe reduces off-target Friedman interactions and, on the higher-order task, lowers RMSE from CV-MLP's 0.1376 to 0.0797 while recovering all nine true pairwise projections. Bike Sharing comparisons and paired RV/CV ablations demonstrate effective use of temporal, calendar, and seasonal context. Noise-performance AUC and learning curves show that the predictive benefits extend to perturbed inputs and limited supervision. Together, these results establish product-unit residual networks as an effective approach to learning and examining nonlinear feature interactions in scientific regression.

Future work will focus on recovering complete higher-order feature sets, improved numerical stabilization for complex product units, and broader robustness evaluation under distribution shift and missing-feature settings.

\section*{CRediT authorship contribution statement}
\textbf{Ziyuan Li:} Conceptualization, Methodology, Software, Data curation, Investigation, Formal analysis, Validation, Visualization, Writing -- original draft,
Writing -- review \& editing.
\textbf{Uwe Jaekel:} Conceptualization, Methodology, Formal analysis, Validation, Writing -- original draft, Writing -- review \& editing.
\textbf{Babette Dellen:} Conceptualization, Methodology, Formal analysis, Validation, Writing -- original draft, Writing -- review \& editing.

\section*{Declaration of competing interest}
The authors declare no competing financial interests or personal relationships that could have influenced the work reported in this paper.

\section*{Data availability}
The datasets are publicly available from the cited sources; synthetic
functions are specified in Eqs.~\eqref{eq:pairwise}, \eqref{eq:higher},
and~\eqref{eq:friedman}. Architecture implementations are available at
\url{https://github.com/ziyuan-li/1d_pure}.

\section*{Declaration of generative AI use}
During the preparation of this manuscript, the authors used OpenAI ChatGPT to assist with language editing, manuscript organization, and literature retrieval. After using this tool, the authors reviewed and edited the content as needed and take full responsibility for the content of the published article.

\bibliographystyle{cas-model2-names}
\bibliography{references}

@article{dellen2024predicting,
  title={Predicting nuclear masses with product-unit networks},
  author={Dellen, Babette and Jaekel, Uwe and Freitas, Paulo SA and Clark, John W},
  journal={Physics Letters B},
  volume={852},
  pages={138608},
  year={2024},
  publisher={Elsevier}
}

@inproceedings{dellen2019function,
  title={Function and pattern extrapolation with product-unit networks},
  author={Dellen, Babette and Jaekel, Uwe and Wolnitza, Marcell},
  booktitle={Computational Science--ICCS 2019: 19th International Conference, Faro, Portugal, June 12--14, 2019, Proceedings, Part II 19},
  pages={174--188},
  year={2019},
  organization={Springer}
}

@inproceedings{li2024data,
  title={Data-Driven {3D} Shape Completion with Product Units},
  author={Li, Ziyuan and Jaekel, Uwe and Dellen, Babette},
  booktitle={International Conference on Computational Science},
  pages={302--315},
  year={2024},
  organization={Springer}
}

@article{leerink1994learning,
  title={Learning with product units},
  author={Leerink, Laurens and Giles, C and Horne, Bill and Jabri, Marwan},
  journal={Advances in neural information processing systems},
  volume={7},
  year={1994}
}

@article{cybenko1989approximation,
  title={Approximation by superpositions of a sigmoidal function},
  author={Cybenko, George},
  journal={Mathematics of control, signals and systems},
  volume={2},
  number={4},
  pages={303--314},
  year={1989},
  publisher={Springer}
}

@inproceedings{molnar2020interpretable,
  title={Interpretable machine learning--a brief history, state-of-the-art and challenges},
  author={Molnar, Christoph and Casalicchio, Giuseppe and Bischl, Bernd},
  booktitle={Joint European conference on machine learning and knowledge discovery in databases},
  pages={417--431},
  year={2020},
  organization={Springer}
}

@article{lundberg2017unified,
  title={A unified approach to interpreting model predictions},
  author={Lundberg, Scott M and Lee, Su-In},
  journal={Advances in neural information processing systems},
  volume={30},
  year={2017}
}

@inproceedings{li2025advancing,
  title={Advancing Complex-Valued Neural Networks with Product Units for {MRI} Reconstruction},
  author={Li, Ziyuan and Jaekel, Uwe and Dellen, Babette},
  booktitle={International Conference on Neural Information Processing},
  pages={540--554},
  year={2025},
  organization={Springer}
}

@inproceedings{he2016deep,
  title={Deep residual learning for image recognition},
  author={He, Kaiming and Zhang, Xiangyu and Ren, Shaoqing and Sun, Jian},
  booktitle={Proceedings of the IEEE conference on computer vision and pattern recognition},
  pages={770--778},
  year={2016}
}

@article{veit2016residual,
  title={Residual networks behave like ensembles of relatively shallow networks},
  author={Veit, Andreas and Wilber, Michael J and Belongie, Serge},
  journal={Advances in neural information processing systems},
  volume={29},
  year={2016}
}

@article{allen2019convergence,
  title={On the convergence rate of training recurrent neural networks},
  author={Allen-Zhu, Zeyuan and Li, Yuanzhi and Song, Zhao},
  journal={Advances in neural information processing systems},
  volume={32},
  year={2019}
}

@inproceedings{tsang2017detecting,
  author    = {Tsang, Michael and Cheng, Dehua and Liu, Yan},
  title     = {Detecting Statistical Interactions from Neural Network Weights},
  booktitle = {International Conference on Learning Representations},
  year      = {2018},
}

@article{durbin1989product,
  title={Product units: A computationally powerful and biologically plausible extension to backpropagation networks},
  author={Durbin, Richard and Rumelhart, David E},
  journal={Neural computation},
  volume={1},
  number={1},
  pages={133--142},
  year={1989},
  publisher={MIT Press One Rogers Street, Cambridge, MA 02142-1209, USA journals-info~…}
}

@article{friedman1991multivariate,
  title={Multivariate adaptive regression splines},
  author={Friedman, Jerome H},
  journal={The annals of statistics},
  volume={19},
  number={1},
  pages={1--67},
  year={1991},
  publisher={Institute of Mathematical Statistics}
}

@article{yeh1998modeling,
  title={Modeling of strength of high-performance concrete using artificial neural networks},
  author={Yeh, I-C},
  journal={Cement and Concrete research},
  volume={28},
  number={12},
  pages={1797--1808},
  year={1998},
  publisher={Elsevier}
}

@article{pace1997sparse,
  title={Sparse spatial autoregressions},
  author={Pace, R Kelley and Barry, Ronald},
  journal={Statistics \& Probability Letters},
  volume={33},
  number={3},
  pages={291--297},
  year={1997},
  publisher={Elsevier}
}

@book{bishop2006pattern,
  title={Pattern recognition and machine learning},
  author={Bishop, Christopher M},
  year={2006},
  publisher={Springer}
}

@article{rumelhart1986learning,
  title={Learning representations by back-propagating errors},
  author={Rumelhart, David E and Hinton, Geoffrey E and Williams, Ronald J},
  journal={nature},
  volume={323},
  number={6088},
  pages={533--536},
  year={1986},
  publisher={Nature Publishing Group UK London}
}

@inproceedings{li2026modeling,
  title={Modeling Nonlinear Feature Interactions with Product-Unit Residual Networks},
  author={Li, Ziyuan and Jaekel, Uwe and Dellen, Babette},
  booktitle={International Conference on Computational Science},
  pages={80--94},
  year={2026},
  organization={Springer}
}

@article{li2025pure,
  title={{PURe}: A Plug-and-Play Product-Unit Residual Module for Vision Networks},
  author={Li, Ziyuan and Jaekel, Uwe and Dellen, Babette},
  journal={arXiv preprint arXiv:2505.04397},
  year={2025}
}

@inproceedings{trabelsi2017deep,
  author    = {Trabelsi, Chiheb and Bilaniuk, Olexa and Zhang, Ying and Serdyuk, Dmitriy and Subramanian, Sandeep and Santos, Jo{\~a}o Felipe and Mehri, Soroush and Rostamzadeh, Negar and Bengio, Yoshua and Pal, Christopher J.},
  title     = {Deep Complex Networks},
  booktitle = {International Conference on Learning Representations},
  year      = {2018},
}

@inproceedings{wang2021dcn,
  title={{DCN V2}: Improved deep \& cross network and practical lessons for web-scale learning to rank systems},
  author={Wang, Ruoxi and Shivanna, Rakesh and Cheng, Derek and Jain, Sagar and Lin, Dong and Hong, Lichan and Chi, Ed},
  booktitle={Proceedings of the web conference 2021},
  pages={1785--1797},
  year={2021}
}

@article{tsang2020does,
  title={How does this interaction affect me? interpretable attribution for feature interactions},
  author={Tsang, Michael and Rambhatla, Sirisha and Liu, Yan},
  journal={Advances in neural information processing systems},
  volume={33},
  pages={6147--6159},
  year={2020}
}

@inproceedings{kingma2014adam,
  author    = {Kingma, Diederik P. and Ba, Jimmy},
  title     = {{Adam}: A Method for Stochastic Optimization},
  booktitle = {International Conference on Learning Representations},
  year      = {2015}
}

@misc{yacht_hydrodynamics_243,
  author       = {Gerritsma, J. and Onnink, R. and Versluis, A.},
  title        = {{Yacht Hydrodynamics}},
  year         = {1981},
  howpublished = {UCI Machine Learning Repository},
  note         = {{DOI}: https://doi.org/10.24432/C5XG7R}
}

@misc{auto_mpg_9,
  author       = {Quinlan, R.},
  title        = {{Auto MPG}},
  year         = {1993},
  howpublished = {UCI Machine Learning Repository},
  note         = {{DOI}: https://doi.org/10.24432/C5859H}
}

@article{fanaee2014event,
  title={Event labeling combining ensemble detectors and background knowledge},
  author={Fanaee-T, Hadi and Gama, Joao},
  journal={Progress in Artificial Intelligence},
  volume={2},
  number={2},
  pages={113--127},
  year={2014},
  publisher={Springer}
}

@misc{bike_sharing_275,
  author       = {Fanaee-T, Hadi},
  title        = {{Bike Sharing}},
  year         = {2013},
  howpublished = {UCI Machine Learning Repository},
  note         = {{DOI}: https://doi.org/10.24432/C5W894}
}

@article{tsanas2009accurate,
  title={Accurate Telemonitoring of {Parkinson's} Disease Progression by Noninvasive Speech Tests},
  author={Tsanas, Athanasios and Little, Max A. and McSharry, Patrick E. and Ramig, Lorraine O.},
  journal={IEEE Transactions on Biomedical Engineering},
  volume={57},
  number={4},
  pages={884--893},
  year={2010},
  doi={10.1109/TBME.2009.2036000}
}

@misc{parkinsons_telemonitoring_189,
  author       = {Tsanas, Athanasios and Little, Max},
  title        = {{Parkinsons Telemonitoring}},
  year         = {2009},
  howpublished = {UCI Machine Learning Repository},
  note         = {{DOI}: https://doi.org/10.24432/C5ZS3N}
}

@article{tsanas2012accurate,
  title={Accurate quantitative estimation of energy performance of residential buildings using statistical machine learning tools},
  author={Tsanas, Athanasios and Xifara, Angeliki},
  journal={Energy and buildings},
  volume={49},
  pages={560--567},
  year={2012},
  publisher={Elsevier}
}

@inproceedings{li2026product,
  title={Product units in gated recurrent units improve nuclear-mass prediction},
  author={Li, Ziyuan and Freitas, Paulo SA and Clark, John W and Dellen, Babette},
  booktitle={International Conference on Computational Science},
  pages={141--156},
  year={2026},
  organization={Springer}
}

@article{li2026pu,
  title={{PU-UNet}: Stable Multiplicative Interactions for Medical Image Segmentation},
  author={Li, Ziyuan and Sufyan, Osamah and Jaekel, Uwe and Dellen, Babette},
  journal={arXiv preprint arXiv:2606.20035},
  year={2026}
}

@inproceedings{virtue2017better,
  title={Better than real: Complex-valued neural nets for {MRI} fingerprinting},
  author={Virtue, Patrick and Yu, Stella X and Lustig, Michael},
  booktitle={2017 IEEE international conference on image processing (ICIP)},
  pages={3953--3957},
  year={2017},
  organization={IEEE}
}

@inproceedings{rendle2010factorization,
  title={Factorization machines},
  author={Rendle, Steffen},
  booktitle={2010 IEEE International conference on data mining},
  pages={995--1000},
  year={2010},
  organization={IEEE}
}

@inproceedings{cheng2020adaptive,
  title={Adaptive factorization network: Learning adaptive-order feature interactions},
  author={Cheng, Weiyu and Shen, Yanyan and Huang, Linpeng},
  booktitle={Proceedings of the AAAI Conference on Artificial Intelligence},
  volume={34},
  pages={3609--3616},
  year={2020}
}

@inproceedings{hendrycks2019using,
  title={Using pre-training can improve model robustness and uncertainty},
  author={Hendrycks, Dan and Lee, Kimin and Mazeika, Mantas},
  booktitle={International conference on machine learning},
  pages={2712--2721},
  year={2019},
  organization={PMLR}
}

@article{janizek2021explaining,
  title={Explaining explanations: Axiomatic feature interactions for deep networks},
  author={Janizek, Joseph D and Sturmfels, Pascal and Lee, Su-In},
  journal={Journal of Machine Learning Research},
  volume={22},
  number={104},
  pages={1--54},
  year={2021}
}

@inproceedings{jayakumar2020multiplicative,
  title={Multiplicative interactions and where to find them},
  author={Jayakumar, Siddhant M and Czarnecki, Wojciech M and Menick, Jacob and Schwarz, Jonathan and Rae, Jack and Osindero, Simon and Teh, Yee Whye and Harley, Tim and Pascanu, Razvan},
  booktitle={International conference on learning representations},
  year={2020}
}

@article{yang2021gami,
  title={{GAMI-Net}: An explainable neural network based on generalized additive models with structured interactions},
  author={Yang, Zebin and Zhang, Aijun and Sudjianto, Agus},
  journal={Pattern Recognition},
  volume={120},
  pages={108192},
  year={2021},
  publisher={Elsevier}
}

@inproceedings{sundararajan2020shapley,
  title={The shapley taylor interaction index},
  author={Sundararajan, Mukund and Dhamdhere, Kedar and Agarwal, Ashish},
  booktitle={International conference on machine learning},
  pages={9259--9268},
  year={2020},
  organization={PMLR}
}

@article{hendrycks2019benchmarking,
  title={Benchmarking Neural Network Robustness to Common Corruptions and Perturbations},
  author={Dan Hendrycks and Thomas Dietterich},
  journal={Proceedings of the International Conference on Learning Representations},
  year={2019}
}

@article{mohr2024learning,
  title={Learning curves for decision making in supervised machine learning: a survey},
  author={Mohr, Felix and van Rijn, Jan N},
  journal={Machine Learning},
  volume={113},
  number={11},
  pages={8371--8425},
  year={2024},
  publisher={Springer}
}

@article{dehghani2021efficiency,
  title={The efficiency misnomer},
  author={Dehghani, Mostafa and Arnab, Anurag and Beyer, Lucas and Vaswani, Ashish and Tay, Yi},
  journal={Proceedings of the International Conference on Learning Representations},
  year={2022}
}

@article{friedman2008predictive,
  title={Predictive learning via rule ensembles},
  author={Friedman, Jerome H. and Popescu, Bogdan E.},
  journal={The Annals of Applied Statistics},
  volume={2},
  number={3},
  pages={916--954},
  year={2008},
}

@article{gu2026interpretable,
  title={Interpretable machine learning for shared feature identification and time series prediction},
  author={Gu, Yuanlin and Zhang, Mao and Li, Baihua and Meng, Qinggang},
  journal={Journal of Computational Science},
  pages={102936},
  year={2026},
  publisher={Elsevier}
}

@book{tibshirani1993introduction,
  title={An Introduction to the Bootstrap},
  author={Efron, Bradley and Tibshirani, Robert J.},
  year={1993},
  publisher={Chapman \& Hall},
  address={New York}
}
\end{document}